\documentclass[letterpaper, 10 pt, conference]{ieeeconf}  % Comment this line out if you need a4paper

\IEEEoverridecommandlockouts                              % This command is only needed if 
\title{\LARGE \bf
FG-CLTP: Fine-Grained Contrastive Language Tactile Pretraining for Robotic Manipulation}

\author{
    Wenxuan Ma$^{1}$, Chaofan Zhang$^{1}$, Yinghao Cai$^{1}$,
    Guocai Yao$^{2}$, Shaowei Cui$^{1,2,*}$, and Shuo Wang$^{1}$     \vspace{2.5mm}\\ % 后三位作者，并增加间距
    $^{1}$Institute of Automation, Chinese Academy of Sciences \\
    $^{2}$Beijing Academy of Artificial Intelligence \\ % 机构列表
    \vspace{-5mm}
    \thanks{$^{*}$Corresponding author: {\tt\small shaowei.cui@ia.ac.cn}
}
}

\usepackage[table]{xcolor}
\usepackage{amssymb}
\usepackage{amsmath,amsfonts}
\usepackage{algorithmic}
\usepackage{algorithm}
\usepackage{array}
\usepackage[strings]{underscore}
\usepackage{textcomp}
\usepackage{stfloats}
\usepackage{url}
\usepackage{verbatim}
\usepackage{graphicx}
\usepackage{cite}
\usepackage{hyperref}
\usepackage{siunitx}
\usepackage{booktabs}
\usepackage{multirow}
\usepackage{pifont}
\usepackage{arydshln}
\usepackage[table]{xcolor}
\newcommand{\cmark}{\textcolor{green}{\checkmark}} 
\newcommand{\xmark}{\textcolor{red}{$\times$}}
\usepackage{graphicx}
\usepackage{cite}
\usepackage{hyperref}
\begin{document}
\maketitle
% \thispagestyle{empty}
% \pagestyle{empty}
% \author{Anonymous Authors}
%%%%%%%%%%%%%%%%%%%%%%%%%%%%%%%%%%%%%%%%%%%%%%%%%%%%%%%%%%%%%%%%%%%%%%%%%%%%%%%%
\begin{abstract}

Recent advancements in integrating tactile sensing into vision-language-action (VLA) models have demonstrated transformative potential for robotic perception. However, existing tactile representations predominantly rely on qualitative descriptors (e.g., texture), neglecting quantitative contact states such as force magnitude, contact geometry, and principal axis orientation, which are indispensable for fine-grained manipulation. To bridge this gap, we propose FG-CLTP, a fine-grained contrastive language tactile pretraining framework. We first introduce a novel dataset comprising over 100k tactile 3D point cloud-language pairs that explicitly capture multidimensional contact states from the sensor’s perspective. We then implement a discretized numerical tokenization mechanism to achieve quantitative-semantic alignment, effectively injecting explicit physical metrics into the multimodal feature space. The proposed FG-CLTP model yields a 95.9\% classification accuracy and reduces the regression error (MAE) by 52.6\% compared to state-of-the-art methods. Furthermore, the integration of 3D point cloud representations establishes a sensor-agnostic foundation with a minimal sim-to-real gap of 3.5\%. Building upon this fine-grained representation, we develop a 3D tactile-language-action (3D-TLA) architecture driven by a flow matching policy to enable multimodal reasoning and control. Extensive experiments demonstrate that our framework significantly outperforms strong baselines in contact-rich manipulation tasks, providing a robust and generalizable foundation for tactile-language-action models.
\end{abstract}

% \begin{IEEEkeywords}
% Tactile Perception, Representation Learning, Vision-Language-Action Models, Contact-Rich Manipulation
% \end{IEEEkeywords}
%%%%%%%%%%%%%%%%%%%%%%%%%%%%%%%%%%%%%%%%%%%%%%%%%%%%%%%%%%%%%%%%%%%%%%%%%%%%%%%%
\section{INTRODUCTION}

Tactile sensing serves as a cornerstone for robust robotic manipulation, particularly in unstructured environments where visual occlusion and complex contact dynamics are prevalent \cite{ye2026visual}. Unlike vision, which typically provides global geometric context, tactile perception offers direct, high-frequency measurements of local contact states, ranging from shear forces and slip detection to micro-texture identification, which are indispensable for tasks requiring high-precision interaction, such as in-hand manipulation and delicate assembly\cite{liuvtdexmanip}. In biological systems, human manipulation dexterity relies heavily on a sophisticated integration of visual planning and tactile feedback, allowing for real-time modulation of grasp force and pose adaptation. However, conventional tactile learning approaches remain largely confined to narrow, task-specific skills, struggling to successfully scale and generalize across diverse manipulation scenarios.

% In recent years, the integration of tactile sensing into the paradigm of foundation models has garnered significant attention. Inspired by the transformative success of vision-language-action (VLA) models\cite{pi2025pi05}, researchers have sought to align tactile data with natural language to leverage the general semantic reasoning capabilities of large language models (LLMs). Pioneering works including UniTouch\cite{yang2024binding}, TVL\cite{fu2024touch}, and CLTP\cite{ma2025cltp}, have successfully demonstrated that tactile representations can be projected into a shared embedding space with vision and language through contrastive learning. By anchoring tactile signals to pre-trained encoders or generating descriptive captions, these models enable robots to articulate surface properties (e.g., rough, sharp, deformable) and reason about object categories. Large-scale pre-training \cite{feng2025anytouch} initiatives have further extended this capability to dynamic temporal sequences, allowing robots to interpret motion-related tactile information across diverse sensor modalities.

In recent years, the integration of tactile sensing into the paradigm of foundation models has garnered significant attention. Inspired by the transformative success of vision-language-models (VLMs) \cite{radford2021learning}, researchers have sought to align tactile data with natural language to leverage the general semantic reasoning capabilities of large language models (LLMs). Early pioneering works such as UniTouch\cite{yang2024binding} and TVL\cite{fu2024touch} successfully demonstrated that 2D tactile images can be projected into a shared embedding space with vision and language. Recent works\cite{feng2025anytouch} further extend this paradigm to dynamic temporal sequences. However, 2D tactile images are typically sensor-specific, as they entangle contact geometry with internal illumination patterns, which complicates cross-sensor generalization. In contrast, 3D point clouds explicitly capture spatial deformation while omitting hardware-specific artifacts, providing a sensor-agnostic representation that generalizes across devices. Building on this property, CLTP\cite{ma2025cltp} aligns 3D tactile point clouds with language through contrastive learning, integrating tactile representation into a large language model for multimodal reasoning. 

\begin{figure}[tbp]
\centering
\includegraphics[width=\linewidth]{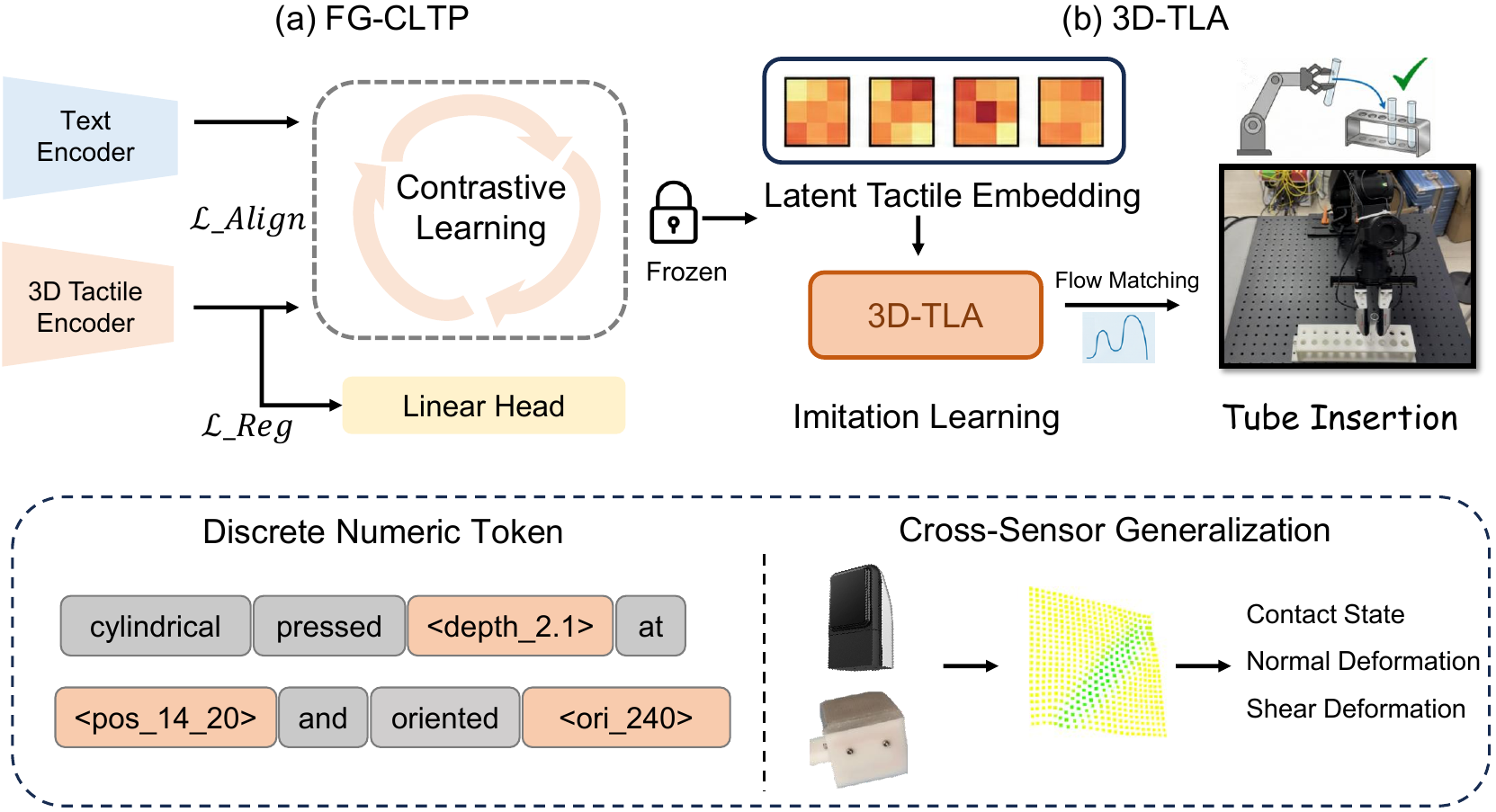}
\caption{Overview of the \textbf{FG-CLTP} Framework. The fine-grained contrastive language-tactile pretraining (FG-CLTP) method aligns 3D tactile point clouds with quantitative contact-state-aware language descriptions. The pretrained encoder is integrated into a flow matching policy (3D-TLA), enabling tactile-based multimodal reasoning and action generation.}
\label{fig:intro}
\end{figure}

Despite these semantic advancements, a critical gap persists between current tactile-language representations and the stringent requirements of fine-grained manipulation. The primary limitation lies in the qualitative nature of existing alignments, which we identify as a lexical bottleneck. While standard linguistic tokens can effectively describe categorical attributes (e.g., recognizing a hard press), they fail to capture the continuous, high-precision physical parameters essential for control, such as distinguishing whether a contact force is 5~N or 20~N, or quantifying the precise penetration depth in millimeters. This lack of quantitative-semantic alignment severs the link between high-level reasoning and low-level execution. As noted in recent action-centric studies \cite{zhang2026clap}, vague semantic adjectives are insufficient for governing the high-frequency, continuous action spaces required for contact-rich tasks. Consequently, existing policies often struggle to generalize from semantic understanding to precise physical execution, leading to suboptimal performance in tasks demanding exact force modulation.

To bridge this gap, we propose \textbf{FG-CLTP} (fine-grained Contrastive language-tactile pretraining), a framework designed to elevate tactile perception from qualitative description to quantitative control. Departing from previous paradigms that rely solely on contrastive alignment of texture and semantics \cite{ma2025cltp}, FG-CLTP introduces a novel quantitative-semantic representation. We first construct a comprehensive dataset of over 100k tactile-language pairs and implement a numerical tokenization strategy. This mechanism explicitly injects physical metrics (e.g., force magnitude, contact location, principal axis) into the language latent space, effectively allowing the model to speak the language of physics. We then learn a semantically aligned and physically grounded tactile representation through multimodal contrastive learning. Furthermore, we incorporate an auxiliary regression loss to learn a continuous embedding space. Building upon this representation, we further develop a 3D tactile-language-action (3D-TLA) architecture built upon a state-of-the-art VLA model, extending its perceptual and policy learning capabilities with physically grounded tactile representations for dexterous manipulation.

We validate the efficacy of our framework through comprehensive benchmarks, ranging from contact state understanding to real-world manipulation tasks. Experimental results demonstrate that FG-CLTP not only achieves superior accuracy in discriminating subtle contact states but also yields statistically significant improvements in policy success rates compared to baselines. 

Our main contributions are summarized as follows:

% \begin{itemize}
    % \item We propose \textbf{FG-CLTP}, a novel framework that that aligns 3D tactile point cloud with quantitative contact-state-aware language description.
\begin{itemize}
    \item We propose \textbf{FG-CLTP}, a contrastive tactile pretraining framework that introduces a discrete numerical tokenization strategy to bridge the gap between qualitative linguistic semantics and quantitative physical contact states. The unified 3D point cloud representation achieves 3.5\% sim2real gap and robust cross-sensor generalization .
    \item We introduce \textbf{Contact3D}, a comprehensive visuo-tactile dataset comprising 100,000 annotated samples across 136 objects, which uniquely pairs 3D deformation point clouds with explicit contact-state labels.
    \item We develop \textbf{3D-TLA}, a downstream flow matching-based policy that seamlessly integrates our tokenized tactile representations in end-to-end contact-rich manipulation tasks.
    % \item We demonstrate through extensive offline benchmarks and real-world experiments that explicitly aligning continuous physical interactions with semantic pretraining is critical for unlocking human-level dexterity in robotic manipulation.
\end{itemize}

\section{Related Work}
\subsection{Tactile Representation Learning}

The heterogeneity of tactile sensors characterized by diverse transduction principles\cite{luo2025tactile} (e.g., optical GelSight\cite{yuan2017gelsight} vs. capacitive arrays) and varying geometries-has long hindered the development of unified and transferable representations \cite{lee2020making}. Early approaches predominantly relied on sensor-specific supervision for tasks such as material classification or slip detection, resulting in limited generalization across hardware platforms. Recent efforts address this limitation through cross-modal alignment, grounding tactile signals in the latent spaces of pre-trained vision-language models (VLMs)\cite{cheng2025touch100k}. UniTouch \cite{yang2024binding} aligns tactile embeddings with CLIP image features to enable zero-shot semantic transfer across sensors. TVL \cite{fu2024touch} extends this paradigm to generative multimodal modeling, while AnyTouch \cite{feng2025anytouch} incorporates temporal dynamics to capture motion-dependent properties. CLTP \cite{ma2025cltp} further aligns tactile 3D point clouds with contact-aware language descriptions and integrates large language models (LLMs) for reasoning and control.

Despite these advances, a severe granularity bottleneck persists: existing methods predominantly align tactile inputs to qualitative descriptors (e.g., rough, hard), leading to coarse semantic grounding. A robot may recognize a force as strong, but cannot infer the precise numerical force magnitude (e.g., 5~N vs. 20~N) necessary for the action adjustments in fine-grained manipulation.

\subsection{Physical Grounding in Vision-Language-Action Models}
% Recent progress attempts to integrate haptic modalities into the vision-language-action (VLA) paradigm. Existing tactile-language-action (TLA) and VTLA architectures typically process tactile inputs by routing tactile images through standard pre-trained image tokenizers, and then feeding them alongside language into foundation models to predict actions. However, this intuitive process introduces a severe domain gap between standard images and tactile images, inherently limiting their performance. VLA-Touch\cite{bi2025vla} utilizes tactile feedback for post-hoc action correction rather than integrating it into action generation. Tactile-VLA\cite{huang2025tactile} and TA-VLA\cite{zhang2025ta} input tactile signals into the VLM backbone via an MLP, but lacks cross-modal semantic alignment, making it difficult for tactile signals to participate in large language model inference. OmniVTLA~\cite{cheng2025omnivtla} integrates a semantically aligned tactile encoder into the VLA model; however, it is confined to coarse, object-level material descriptions (e.g., metal). This qualitative description severely limits the extraction of tactile information such as contact position and force, and lacks the numerical sensitivity required to resolve complex 3D contact states such as deformation depth and shear force. These compounded limitations highlight that to unlock high-precision VTLA, continuous tactile interactions must be translated into a quantitatively and semantically aligned representation to enable the LLM to conduct accurate reasoning and control.

Integrating tactile sensing into vision-language-action (VLA) models has recently gained attention\cite{huang2026tactile}. Most tactile-language-action (TLA)\cite{hao2026tla,yang2025bitla} and VTLA models\cite{zhang2025vtla} process tactile images through standard visual tokenizers and feed them into foundation models for action prediction. However, the domain gap between natural images and tactile imagery limits effective grounding. VLA-Touch \cite{bi2025vla} employs tactile feedback for post-hoc correction rather than direct action generation, while Tactile-VLA \cite{huang2025tactile} and TA-VLA \cite{zhang2025ta} inject tactile signals via lightweight adapters without explicit semantic alignment. OmniVTLA \cite{cheng2025omnivtla} incorporates a semantically aligned tactile encoder but remains confined to coarse object-level descriptions (e.g., material type).

Overall, current VTLA systems largely rely on qualitative tactile semantics, lacking the numerical sensitivity needed to model continuous contact attributes such as force magnitude, contact depth, and shear deformation. This limitation restricts precise physical reasoning and control in contact-rich manipulation.

\subsection{Tokenization Strategies in Multimodal Learning}

Modern multimodal models rely on tokenization, mapping inputs into discrete vocabulary elements for Transformer-based reasoning. While effective for semantic abstraction, standard tokenization strategy struggles to represent precise numerical information essential for robotic control. Recent work explores discretizing continuous perceptual signals into semantic tokens to enhance reasoning. Perception Tokens \cite{bigverdi2025perception} apply VQ-VAE \cite{razavi2019generating} to quantize depth maps into vocabulary elements, enabling structured intermediate predictions. In structured data modeling, TP-BERTa \cite{yan2024making} demonstrates that discretizing scalar values into magnitude-aware tokens significantly improves numerical reasoning in LLMs.

Motivated by these findings, we introduce a discrete numeric tokenization strategy for tactile attributes, bridging the gap between fine-grained quantitative representation and language semantics in multimodal learning. Instead of compressing contact state into qualitative language, we discretize continuous contact properties (e.g., force magnitude and deformation depth) into explicit numeric tokens, enabling quantitatively grounded semantic alignment and physically-informed reasoning.

\section{Contact3D Dataset}
\begin{figure*}[t]
\vspace{2mm}
\centering
\includegraphics[width=\textwidth]{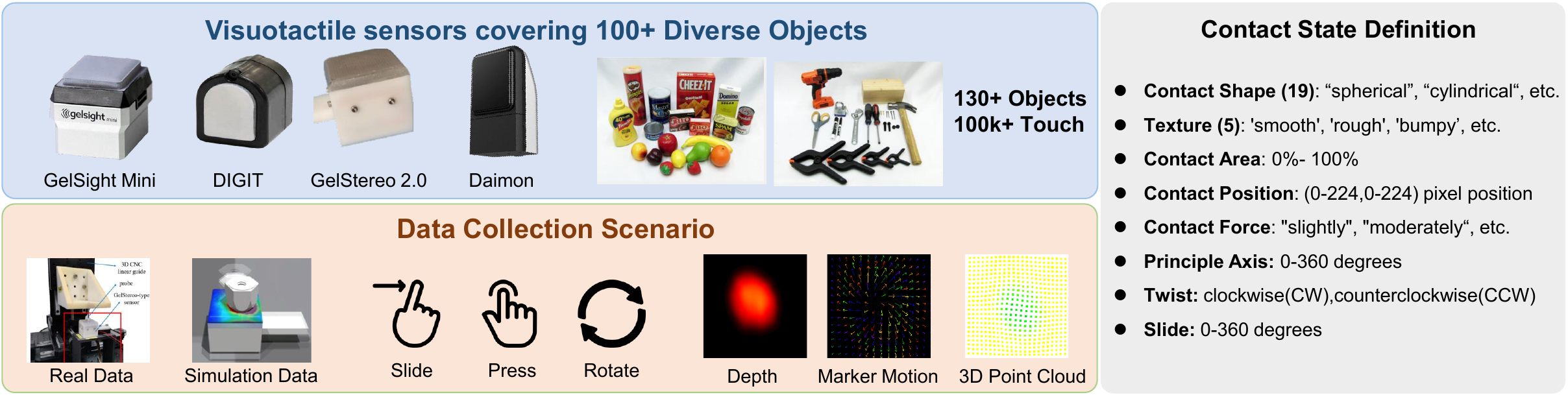}
\caption{Overview of the Contact3D Dataset. The Contact3D dataset integrates real-world and simulated multi-sensor data, implementing automated collection through pressing, sliding, and rotating primitives, alongside comprehensive contact state annotations for holistic representation.}
\label{fig:dataset}
\end{figure*}

We construct Contact3D, a large-scale multimodal dataset designed to support contact-state-aware tactile representation learning for robotic manipulation, as shown in Fig. \ref{fig:dataset}. The dataset comprises a diverse set of 136 objects, including YCB objects, standard industrial components from \href{https://www.mcmaster.com/}{McMaster}, and custom-designed peg geometries. As shown in Tab. \ref{tab:comparison}, unlike prior datasets that primarily emphasize superficial or qualitative tactile attributes, Contact3D is explicitly designed to align tactile signals with fine-grained contact-state descriptions from the perspective of tactile sensing. For each data sample, we collect a synchronized multimodal tuple consisting of tactile 3D point clouds, tactile images, explicit physical signals (force and torque), and detailed contact-state annotations relevant to manipulation tasks, such as deformation shape, contact position and principal axis orientation.

\subsection{Data Collection}
To achieve high-fidelity simulation of soft-body contact dynamics for visuotactile sensors, we adopt TacFlex\cite{zhang2025tacflex}, a visuotactile simulation framework built upon the Isaac Gym FEM simulator. TacFlex enables the simultaneous generation of multimodal tactile signals, including high-resolution tactile images, marker displacement fields, 3D deformation point clouds, and 6-DoF contact force and torque measurements. This unified simulation setup allows us to capture both geometric deformation and physical interaction signals under diverse contact conditions.

We employ a primitives-based data collection strategy to systematically cover common contact interactions encountered in robotic manipulation. Specifically, three fundamental contact primitives including pressing, sliding, and twisting are defined. For each object, multiple contact points are randomly sampled on the object surface, and the tactile sensor is driven toward the object along directions sampled within a $15^\circ$ cone around the local surface normal to establish contact. During interaction, the sensor follows motion trajectories that induce varying contact forces, ranging from light touch to substantial deformation. At each time step, we record the tactile 3D deformation point cloud $\mathbf{P} \in \mathbb{R}^{N \times 3}$, marker displacement fields, and the corresponding 6-DoF force--torque vector $\mathbf{F} \in \mathbb{R}^6$. In addition, we automatically annotate manipulation-relevant contact attributes, including contact location, contact area size, and the principal axis of the contact patch. This data collection process ensures that the resulting dataset captures not only the geometric structure of contact but also the dynamic physical properties that are critical for contact-rich manipulation.
\begin{table}[t]
\centering
\caption{Comparison of different tactile datasets. Our \textbf{Contact3D} dataset provides comprehensive features including depth, force, language and dynamic interactions.}
\label{tab:comparison}
\resizebox{\linewidth}{!}{%
\begin{tabular}{ccccccccc}
\toprule
\textbf{Dataset} & \textbf{Sensor} & \textbf{Objects} & \textbf{Samples} & \textbf{Marker} & \textbf{Depth} & \textbf{Force} & \textbf{Lang.} & \textbf{Dynamic} \\ 
\midrule
VisGel \cite{yuan2017connecting}           & GelSight      & 118         & 3.0k & \cmark & \xmark               & \cmark           & \cmark & \xmark \\
Cloth \cite{yuan2018active}             & GelSight      & 153         & 90k  & \cmark & \xmark               & \cmark             & \cmark & \cmark \\
Touch and Go \cite{yang2022touch} & GelSight      & -  & 13.9k & \cmark & \cmark               & \cmark                   & \cmark & \cmark \\
SSVTP \cite{kerr2022ssvtp}             & DIGIT         & 15          & 4.5k & \xmark & \xmark               & \xmark                   & \xmark & \xmark \\
TVL \cite{fu2024touch}                 & DIGIT         & - & 44k  & \xmark & \xmark               & \xmark                   & \cmark & \xmark \\
Octopi \cite{yu2024octopi}           & GelSight & 74          & 39k  & \xmark & \xmark               & \xmark                   & \cmark & \cmark \\
TacQuad \cite{feng2025anytouch}        & 4 Sensors     & 124         & 72k  & \xmark & \xmark               & \cmark                   & \cmark & \cmark \\
TCL3D \cite{ma2025cltp}      & 3 Sensors     & 117         & 50k  & \xmark & \cmark               & \cmark                   & \cmark & \xmark \\ 
\hdashline
\textbf{Contact3D (Ours)   }                & 4 Sensors     & 136         & 100k  & \cmark & \cmark               & \cmark                   & \cmark & \cmark \\ 
\bottomrule
\end{tabular}%
}
\end{table}
\subsection{Language Labeling}We adopt a hybrid annotation strategy that combines VLM-assisted automatic labeling with analysis-based contact state estimation. For natural language attributes describing 3D deformation shape and surface texture, we employ Qwen3-VL-8B-Instruct\cite{yang2025qwen3} to generate textual labels. Empirically, we find that providing the model with rendered contact depth maps together with a predefined candidate list enables the LLM to produce pseudo-labels that are highly consistent with human judgments. We attribute this to the fact that depth maps convey clearer geometric and shape cues compared to raw RGB observations. In rare cases where the model fails to disambiguate due to ambiguous or irregular deformation patterns, the corresponding samples are conservatively labeled as irregular.

For quantitative contact properties, including contact depth and contact location, we derive labels through analytical computation based on the 3D deformation point cloud and marker displacement fields. The principal deformation axis is estimated via singular value decomposition (SVD) of the deformation point cloud. To characterize sliding and twisting motions, we first extract the contact region according to the estimated contact depth, then compute the mean displacement and local rotational components of the marker displacement field within this region. Sliding direction and principal axis orientation are both mapped to a continuous 360-degree representation, while twisting is discretized into clockwise and counterclockwise categories. Samples with weak or unreliable signals, such as an SVD singularity ratio below 2 or a total rotation amplitude below a predefined threshold, will be marked as invalid values for the corresponding attribute.

Finally, all estimated attributes are assembled into structured natural language contact-state descriptions using predefined prompt templates, e.g.,
“A [texture] [shape] object, pressed [depth] at [position] with a [area] contact area, oriented along [axis], sliding towards [orientation] and twisting [clockwise].”
For each base description, we further apply LLM to perform language augmentation, generating ten semantically equivalent but syntactically diverse descriptions. In summary, we obtain 100k annotated tactile samples.

\section{Method}
We propose a framework for learning a unified tactile representation that bridges abstract semantic understanding and precise physical states, serving as a foundation for contact-rich manipulation tasks. To achieve this goal, a quantitative-semantic alignment strategy is introduced. This strategy augments the text modality with explicit numerical tokens and incorporates an auxiliary regression objective, enabling the encoder to model fine-grained physical quantities while preserving robust semantic abstraction.

\begin{figure*}[t]
\vspace{2mm}
\centering
\includegraphics[width=\textwidth]{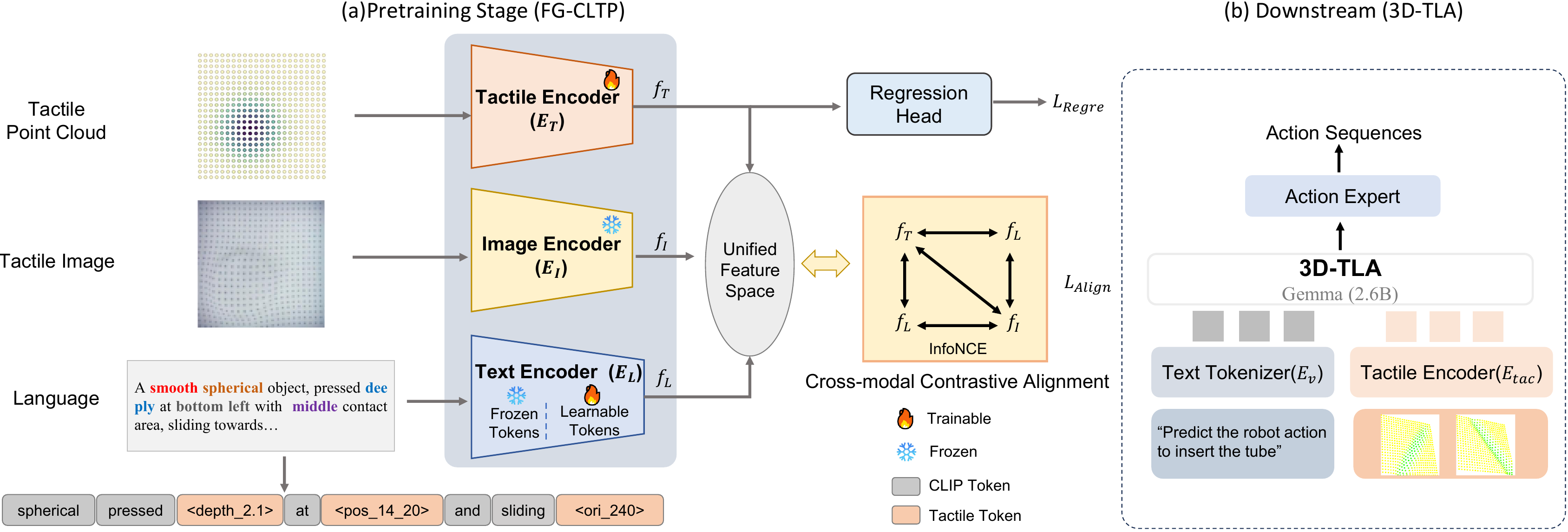}
\caption{Architecture of the FG-CLTP. Tactile point clouds, tactile images, and language descriptions are processed through their respective encoders. For text tokenization, discrete numerical tokens are introduced for fine-grained alignment. The original tokens are frozen, while the newly added tokens are learnable. Contrastive learning is employed for semantic alignment within the feature space. Additionally, an explicit physical attribute regression loss is incorporated to enhance the discriminability of specific physical properties.}
\label{fig:method}
\end{figure*}

\subsection{Unified 3D Tactile Representation}

Unlike unordered point clouds in computer vision, visuotactile signals possess a fixed topological structure corresponding to physical sensor markers. Defining the marker positions in rest and contact states as $\mathcal{M}_{rest} = \{\mathbf{p}_i^0\}_{i=1}^N$ and $\mathcal{M}_{t} = \{\mathbf{p}_i^t\}_{i=1}^N$ respectively, the input $\mathcal{M}_{t}$ implicitly encodes the 3D deformation field $\Delta \mathcal{M}_t = \mathcal{M}_{t} - \mathcal{M}_{rest}$. This unified formulation seamlessly generalizes to any tactile sensor with surface displacement and depth maps.

This contact-state-aware representation explicitly captures normal deformations and tangential shear patterns, providing direct physical cues for complex contact interactions. By processing these structured coordinates, the encoder directly extracts physics-informed features that align with fine-grained semantic descriptions (e.g., twisting clockwise). This paradigm circumvents the need for explicit force calibration while preserving the precise spatial force distributions essential for dexterous manipulation.
\subsection{Discrete Numeric Tokenization}

Traditional multimodal alignment relies on qualitative adjectives (e.g., large, tilted), which lack the precision required for tactile-sensitive manipulation. To bridge the gap between continuous physical states (e.g., depth, orientation) and discrete language tokens, a quantitative numeric tokenization strategy is introduced. Continuous tactile attributes are discretized into bins and mapped to unique tokens added to the vocabulary of the language encoder. For example, contact depth is mapped to tokens ranging from \texttt{<depth_0.0>} to \texttt{<depth_4.0>}, and contact area spans \texttt{<area_0.01>} to \texttt{<area_1.0>}. To enable spatial reasoning over 3D deformations, angle-specific tokens (e.g., \texttt{<principal_0>} to \texttt{<principal_360>}) are introduced to represent the principal axis of deformation, together with position tokens for precise contact localization.

These tokens are embedded into natural language descriptions to form digital-enhanced prompts. Furthermore, the textual descriptions incorporate temporal contact primitives that capture dynamic actions such as sliding and twisting, ensuring that the representation encodes both static geometry and the temporal evolution of contact.

\subsection{Contrastive Language Tactile Pretraining}

The 3D tactile point cloud $T$, the digital-enhanced language description $L$, and the rendered tactile image $I$ are aligned in a unified feature space. The pre-trained CLIP architecture \cite{radford2021learning} is adopted as the backbone, with encoders denoted by $E_T$, $E_L$, and $E_I$, respectively. To accommodate the newly introduced digital tokens without inducing catastrophic forgetting of pre-trained knowledge, a partial fine-tuning strategy is employed. Within $E_L$, the embeddings of the original CLIP vocabulary are frozen, and only the embeddings of the new digital tokens are optimized. This design encourages alignment between tactile features and both broad semantic concepts (frozen) and precise digital states (learnable).

Given a batch of $N$ triplets $(T_i, L_i, I_i)$, features are extracted as $f_T = E_T(T)$, $f_L = E_L(L)$, and $f_I = E_I(I)$. An InfoNCE loss is used to maximize the cosine similarity between matched pairs. The tactile-language alignment loss is defined as
\begin{equation}
\mathcal{L}_{T\rightarrow L} = - \frac{1}{2} \sum_{i=1}^{N} \log \frac{\exp(f_{T_i} \cdot f_{L_i} / \tau)}{\sum_{j=1}^{N} \exp(f_{T_i} \cdot f_{L_j} / \tau)},
\end{equation}
where $\tau$ is a learnable temperature parameter. The total alignment objective aggregates interactions among tactile ($T$), language ($L$), and image ($I$) modalities:
% \begin{equation}
% \mathcal{L}_{Align} = \lambda_{TL}\mathcal{L}_{T\rightarrow L} + \lambda_{TI}\mathcal{L}_{T\rightarrow I} + \lambda_{LI}\mathcal{L}_{L\rightarrow I},
% \end{equation}
\begin{equation}
\begin{aligned}
\mathcal{L}_{Align} &= \frac{\lambda_{TL}}{2}(\mathcal{L}_{T\rightarrow L} + \mathcal{L}_{L\rightarrow T}) + \frac{\lambda_{TI}}{2}(\mathcal{L}_{T\rightarrow I} + \mathcal{L}_{I\rightarrow T}) \\
&+ \frac{\lambda_{LI}}{2}(\mathcal{L}_{L\rightarrow I} + \mathcal{L}_{I\rightarrow L}),
\end{aligned}
\end{equation}
where $\mathcal{L}_{T\rightarrow L}$ denotes the contrastive loss between modalities $T$ and $L$. This optimization ensures that $E_T$ learns a representation that is semantically aligned, physically grounded and numerically sensitive.

\subsection{Auxiliary Physical Regression}

Although contrastive learning aligns global semantics, it may overlook fine-grained local physical details. To address this limitation, an auxiliary regression head is introduced to provide direct numerical supervision for the tactile encoder. Consider a tactile sample $T_i$ associated with a ground-truth physical state vector $\mathbf{y}_i \in \mathbb{R}^V$, where each dimension corresponds to a normalized tactile attribute (e.g., depth, contact position, principal axis angle). A lightweight regression network (MLP) is trained on top of the tactile features to predict these values using mean squared error (MSE) loss:
\begin{equation}
\mathcal{L}_{Regre} = \frac{1}{V} \sum_{c=1}^{V} | \hat{y}_{i,c} - y_{i,c} |^2,
\end{equation}
where $c$ denotes the channel of the physical attribute. This regression loss explicitly enforces the encoding of precise physical quantities. The final training objective is
\begin{equation}
\mathcal{L}_{Total} = \mathcal{L}_{Align} + \mathcal{L}_{Regre}.
\end{equation}
This combined supervision guides the model to capture fine-grained tactile-physical alignments beyond simple semantic associations.

\subsection{3D-TLA Policy Learning}
% \begin{figure}[!t]
% \centering
% \includegraphics[width=\linewidth]{figures/tla.pdf}
% \caption{3D-TLA Framework.}
% \label{fig:tla}
% \end{figure}

We introduce 3D-TLA, a tactile-language-action model built upon the $\pi_{0.5}$ architecture \cite{pi2025pi05}. The framework comprises three primary modules: a multimodal tokenizer suite, a transformer backbone, and a flow matching action head, as shown in Fig.\ref{fig:method}. Specifically, language instructions $l_t$, visual observations $I_t$, and tactile 3D point clouds ($N \times 3$) are respectively tokenized by PaliGemma, SigLIP \cite{zhai2023sigmoid}, and our proposed tactile encoder. These concatenated tokens are fed into a Gemma-2B backbone to extract high-level spatio-temporal features that condition the action generation. The robot's control space is defined as $a_t \in \mathbb{R}^{10}$, comprising 3-DoF translation, 6-DoF rotation, and a binary gripper state. To model the non-linear dynamics of contact-rich manipulation, we employ flow matching. Conditioned on the multimodal embeddings (tactile $z_T$, visual $z_I$, and language $L$), the policy $\pi_\theta$ learns a time-dependent vector field $v_t(\mathbf{x})$ to transform a standard Gaussian distribution $p_0$ into the expert action distribution $p_1$. To prevent instability caused by uninitialized tactile embeddings, training proceeds in two stages: initially freezing the backbone to train only the alignment MLPs, followed by jointly fine-tuning the MLPs and the backbone via LoRA.

\section{Experiments}

We validate the proposed tactile representations across offline benchmarks to demonstrate their quantitative-semantic alignment and numerical sensitivity, and through real-world tasks to confirm that these aligned representations can be effectively integrated into VLA models for learning contact-rich manipulation.

\textbf{Experimental Setup.} We base our model on ULIP-2\cite{xue2024ulip}. For the 3D tactile input, we uniformly sample Np = 529 points. We train and evaluate our model on the Contact3D dataset with a ratio of 9:1. For different tactile sensors, we normalize their scales to a unified spatial range.

\subsection{Offline Benchmark Evaluation}
% We evaluate the quality of the learned tactile representations through a comprehensive suite of offline benchmarks. Specifically, we conduct  supervised linear probing, and continuous attribute regression.

\subsubsection{Contact State Classification}
To assess the fine-grained discriminative capabilities of the learned features, we freeze the pre-trained tactile encoder and train linear probes. We compare our approach against state-of-the-art tactile representation baselines, including TVL~\cite{fu2024touch}, UniTouch~\cite{yang2024binding}, AnyTouch~\cite{feng2025anytouch}, and CLTP~\cite{ma2025cltp}.

\begin{figure}[!t]
\centering
\includegraphics[width=\linewidth]{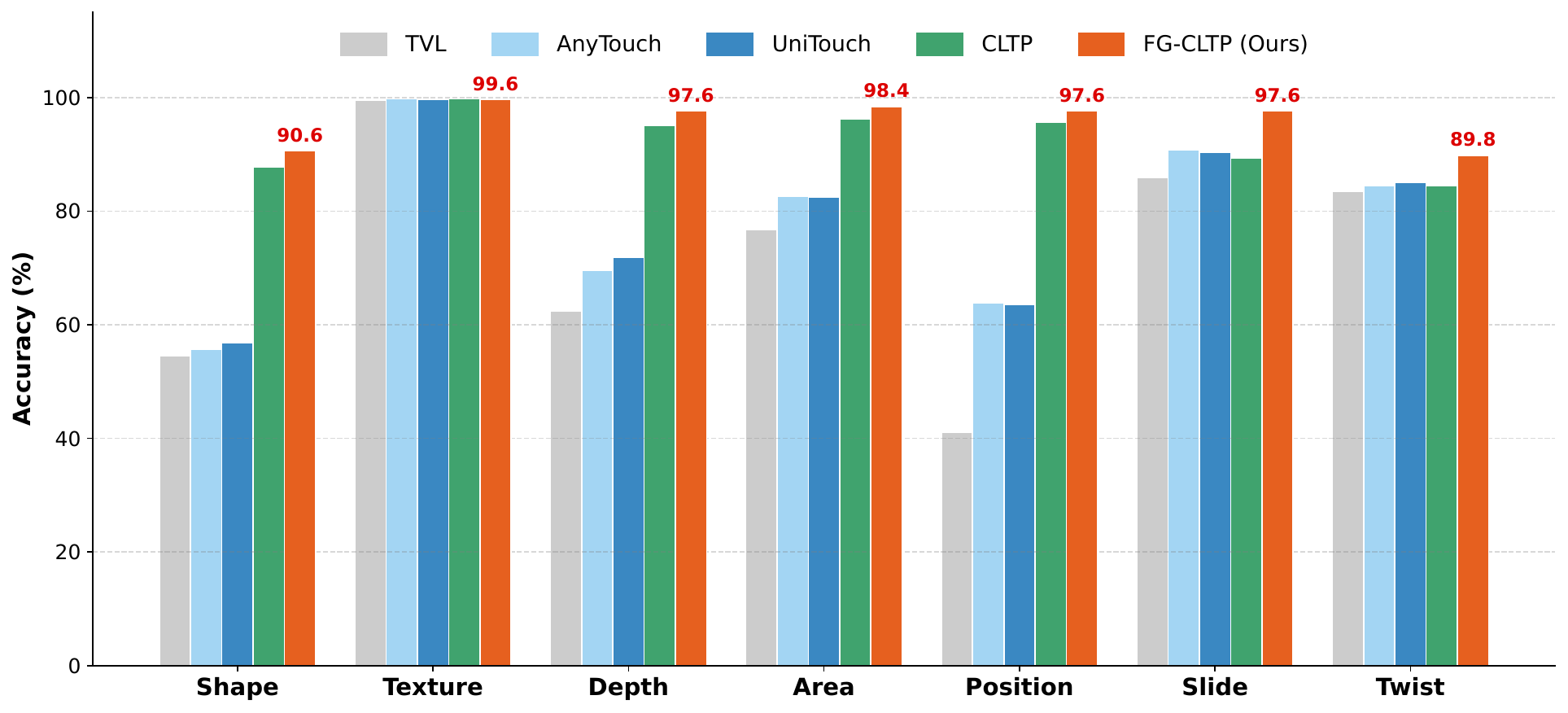}
\caption{Contact State Classification Results.}
\label{fig:cls}
\end{figure}

As illustrated in Fig.~\ref{fig:cls}, our model achieves 90.6\% accuracy in shape classification, and 97.6\% in both depth and position classification. It substantially outperforms vision-based tactile representations, which typically project signals into 2D textural spaces and consequently struggle with precise 3D geometric contact recognition. Crucially, these results confirm that anchoring 3D tactile point clouds to a discretized numerical vocabulary effectively resolves the metric ambiguity inherent in prior continuous embeddings, thereby fundamentally enhancing fine-grained contact state comprehension.

\subsubsection{Contact State Regression}
\begin{table*}[t]
\vspace{2mm}
\centering
\caption{Comparison across multiple regression targets.} 
\label{tab:regression}
\small
\setlength{\tabcolsep}{2.5pt} 
\resizebox{\linewidth}{!}{
    \begin{tabular}{
    l
    S[table-format=1.3] S[table-format=1.3] S[table-format=1.3]
    S[table-format=1.3] S[table-format=1.3] S[table-format=1.3]
    S[table-format=1.3] S[table-format=1.3] S[table-format=1.3]
    S[table-format=1.3] S[table-format=1.3] S[table-format=1.3]
    S[table-format=1.3] S[table-format=1.3] S[table-format=1.3]
    S[table-format=1.3] S[table-format=1.3] S[table-format=1.3]
    }
    \toprule
    \multirow{2}{*}{Method} &
    \multicolumn{3}{c}{Depth} &
    \multicolumn{3}{c}{Position} &
    \multicolumn{3}{c}{Area} &
    \multicolumn{3}{c}{Principal Axis} &
    \multicolumn{3}{c}{Shear} &
    \multicolumn{3}{c}{Macro Avg.} \\
    \cmidrule(lr){2-4}\cmidrule(lr){5-7}\cmidrule(lr){8-10}\cmidrule(lr){11-13}\cmidrule(lr){14-16}\cmidrule(lr){17-19}
    & {MAE~($\downarrow$)} & {RMSE~($\downarrow$)} & {$R^2$~($\uparrow$)}
    & {MAE~($\downarrow$)} & {RMSE~($\downarrow$)} & {$R^2$~($\uparrow$)}
    & {MAE~($\downarrow$)} & {RMSE~($\downarrow$)} & {$R^2$~($\uparrow$)}
    & {MAE~($\downarrow$)} & {RMSE~($\downarrow$)} & {$R^2$~($\uparrow$)}
    & {MAE~($\downarrow$)} & {RMSE~($\downarrow$)} & {$R^2$~($\uparrow$)}
    & {MAE~($\downarrow$)} & {RMSE~($\downarrow$)} & {$R^2$~($\uparrow$)} \\
    \midrule
TVL\cite{fu2024touch} & 0.268 & 0.340 & 0.499 & 0.346 & 0.453 & 0.492 & 0.121 & 0.176 & 0.858 & 0.494 & 0.601 & 0.043 & 0.373 & 0.478 & 0.286 & 0.320 & 0.410 & 0.436 \\
AnyTouch\cite{feng2025anytouch} & 0.219 & 0.283 & 0.650 & 0.252 & 0.353 & 0.694 & 0.084 & 0.118 & 0.937 & 0.401 & 0.521 & 0.275 & 0.206 & 0.302 & 0.718 & 0.232 & 0.315 & 0.655 \\
UniTouch\cite{yang2024binding} & 0.192 & 0.250 & 0.729 & 0.246 & 0.346 & 0.703 & 0.075 & 0.109 & 0.946 & 0.346 & 0.468 & 0.423 & 0.240 & 0.343 & 0.630 & 0.220 & 0.303 & 0.686 \\
CLTP\cite{ma2025cltp} & 0.093 & 0.120 & 0.938 & 0.124 & 0.162 & 0.935 & 0.070 & 0.098 & 0.956 & 0.249 & 0.373 & 0.626 & 0.225 & 0.323 & 0.676 & 0.152 & 0.215 & 0.826 \\
\rowcolor[HTML]{DAEFF2} FG-CLTP (w/o token) & 0.082 & 0.108 & 0.945 & 0.096 & 0.138 & 0.950 & 0.068 & 0.095 & 0.960 & 0.145 & 0.210 & 0.815 & 0.130 & 0.185 & 0.840 & 0.104 & 0.147 & 0.902 \\
\rowcolor[HTML]{DAEFF2} FG-CLTP & \textbf{0.072} & \textbf{0.098} & \textbf{0.958} & \textbf{0.078} & \textbf{0.120} & \textbf{0.964} & \textbf{0.066} & \textbf{0.092} & \textbf{0.964} & \textbf{0.067} & \textbf{0.122} & \textbf{0.959} & \textbf{0.078} & \textbf{0.117} & \textbf{0.955} & \textbf{0.072} & \textbf{0.110} & \textbf{0.960} \\ \hline

    \bottomrule
    \end{tabular}
}
\end{table*}

\begin{figure}[t]
\vspace{2mm}
\centering
\includegraphics[width=\linewidth]{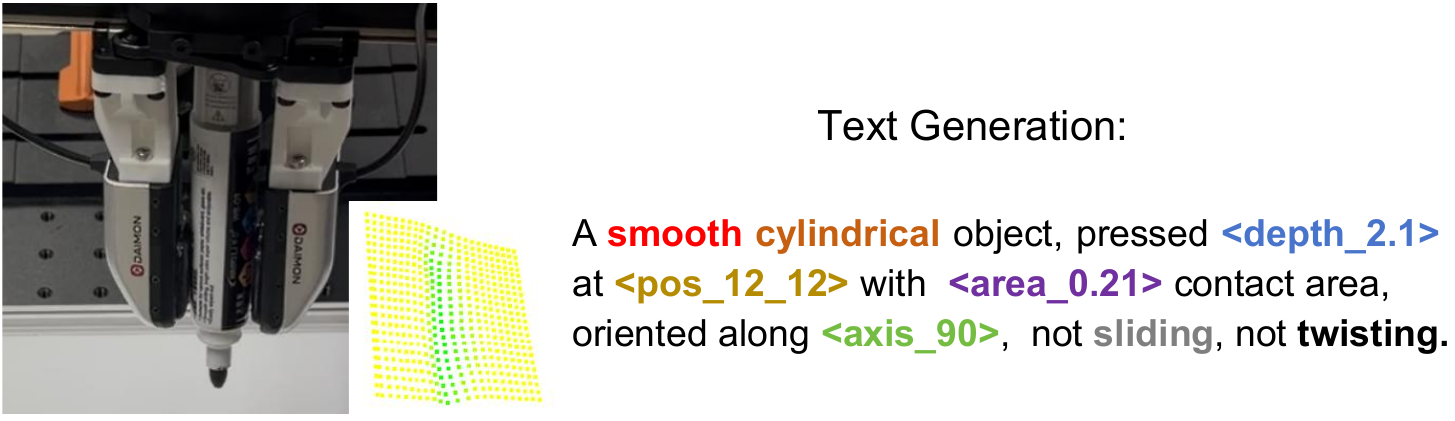}
\caption{A case study of generating text descriptions from realistic tactile point clouds.}
\label{fig:case}
\end{figure}
To further evaluate physical fidelity, we conduct a regression benchmark targeting continuous contact attributes, including contact depth (mm), contact position (pixel), contact area (percent), principal axis angle (degrees), and shear orientation (degrees). A lightweight MLP regressor is attached to the frozen tactile encoder. 

% The proposed FG-CLTP demonstrates a decisive advantage in decoding complex contact states, establishing a state-of-the-art across physical regression benchmarks. As detailed in Table~\ref{tab:regression}, our method achieves a Macro $R^2$ of 0.958, significantly surpassing previous tactile representation baselines such as UniTouch ($0.686$) and AnyTouch ($0.655$). A critical disparity is observed in deformation-sensitive attributes; for instance, in shear force estimation, vision-based methods like TVL struggle with a high MAE of 0.373, whereas our approach significantly minimizes this error to 0.078, while maintaining a robust $R^2$ of 0.955. Compared to the closest competitor CLTP, our method reduces the Macro MAE by approximately 52\% ($0.151 \rightarrow 0.072$), proving that the fine-grained tokenization strategy effectively captures subtle geometric variations that are often lost in coarse-grained contrastive learning. This high-precision regression capability aligns with the classification results, where the model attains 97.6\% accuracy in depth discrimination, further confirming the physical fidelity of the learned embeddings. It is noteworthy that our method significantly outperforms baseline approaches in shear force and slip recognition. This is because our 3D tactile representation effectively captures both tangential and normal surface deformations, providing critical and informative data.

The proposed FG-CLTP demonstrates a decisive advantage in decoding complex contact states, establishing a comprehensive state-of-the-art across physical regression benchmarks. As detailed in Table~\ref{tab:regression}, our method achieves a macro average $R^2$ of 0.960, significantly surpassing previous tactile representation baselines such as UniTouch (0.686) and AnyTouch (0.655). A critical disparity is observed in deformation-sensitive attributes; for instance, in shear force estimation, vision-based methods like TVL struggle with a high MAE of 0.373, whereas our approach significantly minimizes this error to 0.078 while maintaining a robust $R^2$ of 0.955. This superiority extends to other complex geometric properties like the principal axis, where FG-CLTP elevates the $R^2$ to 0.959 compared to the 0.626 achieved by previous methods. Compared to the closest competitor CLTP, our method reduces the macro average MAE by approximately 52.6\% (from 0.152 to 0.072), proving that the fine-grained tokenization strategy effectively aggregates localized features and captures subtle geometric variations that are often lost in coarse-grained contrastive learning. The ablation study further validates this improvement, showing that the token integration substantially refines predictions over the base FG-CLTP (w/o token) configuration across all targets. It is noteworthy that our method significantly outperforms baseline approaches in challenging tasks like shear force recognition because our 3D tactile representation effectively captures both tangential and normal surface deformations. This high-precision regression capability provides critical and informative data that aligns with the classification results, where the model attains 97.6\% accuracy in depth discrimination, further confirming the physical fidelity of the learned embeddings.

\subsubsection{Cross-Sensor Generalization}
Although our model was trained exclusively on the Contact3D dataset, we conducted cross-sensor validation using real-world data collected by the GelStereo 2.0~\cite{zhang2023gelstereo} and DM-Tac M tactile sensors. By aligning all data to a uniform 3D tactile representation, which enables cross-sensor generalization, we evaluate the resulting contact state classification performance.
\begin{table}[htbp]
\vspace{2mm}
\centering
% \usepackage{multirow}
\caption{Cross-Sensor Generalization Results.}
\resizebox{\linewidth}{!}{
\begin{tabular}{cccccccccc}
\hline
\multirow{2}{*}{\textbf{Domain}} & \multirow{2}{*}{\textbf{Method}} & \multicolumn{7}{c}{\textbf{Physical Property Tasks}} & \multirow{2}{*}{\textbf{Avg.}} \\ \cline{3-9}
 &  & Shape & Texture & Depth & Area & Pos. & Slide & Twist &  \\ \hline
\multirow{5}{*}{\textbf{\begin{tabular}[c]{@{}c@{}}GelStereo 2.0 \\ (Simulation)\end{tabular}}} & TVL & 54.6 & 99.5 & 62.4 & 76.7 & 41.1 & 85.9 & 83.5 & 72.0 \\
 & AnyTouch & 55.7 & 99.8 & 69.6 & 82.6 & 63.8 & 90.7 & 84.4 & 78.1 \\
 & UniTouch & 56.8 & 99.6 & 71.9 & 82.4 & 63.5 & 90.3 & 85.1 & 78.5 \\
 & CLTP & 87.7 & 99.8 & 95.1 & 96.2 & 95.7 & 89.4 & 84.4 & 92.6 \\
 & \cellcolor[HTML]{DAEFF2}\textbf{FG-CLTP (Ours)} & \cellcolor[HTML]{DAEFF2}\textbf{90.6} & \cellcolor[HTML]{DAEFF2}\textbf{99.8} & \cellcolor[HTML]{DAEFF2}\textbf{97.6} & \cellcolor[HTML]{DAEFF2}\textbf{98.4} & \cellcolor[HTML]{DAEFF2}\textbf{97.6} & \cellcolor[HTML]{DAEFF2}\textbf{97.6} & \cellcolor[HTML]{DAEFF2}\textbf{89.8} & \cellcolor[HTML]{DAEFF2}\textbf{95.9} \\ \hline
\multirow{5}{*}{\textbf{\begin{tabular}[c]{@{}c@{}}GelStereo 2.0\\ (Real-World)\end{tabular}}} & TVL & 48.2 & 95.4 & 55.8 & 70.1 & 35.4 & 78.2 & 75.6 & 65.5 \\
 & AnyTouch & 50.5 & 96.1 & 62.3 & 76.5 & 58.2 & 84.5 & 78.9 & 72.4 \\
 & UniTouch & 52.1 & 96.5 & 65.8 & 77.2 & 59.4 & 85.1 & 79.4 & 73.6 \\
 & CLTP & 83.4 & 97.2 & 89.5 & 92.8 & 90.1 & 85.6 & 80.2 & 88.4 \\
 & \cellcolor[HTML]{DAEFF2}\textbf{FG-CLTP (Ours)} & \cellcolor[HTML]{DAEFF2}\textbf{87.2} & \cellcolor[HTML]{DAEFF2}\textbf{98.1} & \cellcolor[HTML]{DAEFF2}\textbf{93.4} & \cellcolor[HTML]{DAEFF2}\textbf{95.6} & \cellcolor[HTML]{DAEFF2}\textbf{93.8} & \cellcolor[HTML]{DAEFF2}\textbf{92.5} & \cellcolor[HTML]{DAEFF2}\textbf{86.3} & \cellcolor[HTML]{DAEFF2}\textbf{92.4} \\ \hline
\multirow{2}{*}{\textbf{\begin{tabular}[c]{@{}c@{}}DM-Tac\\ (Generalization)\end{tabular}}}
 & CLTP & 75.8 & 94.6 & 81.2 & 82.6 & 80.4 & 79.8 & 76.1 & 81.5 \\
 & \cellcolor[HTML]{DAEFF2}\textbf{FG-CLTP (Ours)} & \cellcolor[HTML]{DAEFF2}\textbf{81.5} & \cellcolor[HTML]{DAEFF2}\textbf{96.8} & \cellcolor[HTML]{DAEFF2}\textbf{86.5} & \cellcolor[HTML]{DAEFF2}\textbf{89.2} & \cellcolor[HTML]{DAEFF2}\textbf{84.3} & \cellcolor[HTML]{DAEFF2}\textbf{85.4} & \cellcolor[HTML]{DAEFF2}\textbf{81.7} & \cellcolor[HTML]{DAEFF2}\textbf{86.5} \\ \hline
\end{tabular}}
\end{table}
Experimental results demonstrate that the proposed FG-CLTP framework effectively bridges the sim-to-real gap. For the GelStereo 2.0 sensor, FG-CLTP achieves an average accuracy of 95.9\% on simulated data and maintains a high accuracy of 92.4\% during real-world testing. This minimal performance drop of only 3.5\% represents the smallest sim-to-real gap among all evaluated methods, significantly outperforming prior architectures such as UniTouch and AnyTouch, which exhibit larger drops of 4.9\% and 5.7\%, respectively. The cross-domain stability of FG-CLTP validates that aligning the data to a unified 3D representation successfully mitigates domain-specific biases, allowing the model—trained exclusively on Contact3D simulation data—to be robustly deployed in the real world.

Beyond direct sim-to-real transfer on the same sensor type, FG-CLTP exhibits strong zero-shot generalization to the unseen DM-Tac sensor, achieving an average accuracy of 86.5\% and consistently outperforming all baseline models. Despite an expected performance decline compared to the GelStereo 2.0 real-world results due to inherent differences in sensor surface characteristics and elastomer profiles, FG-CLTP maintains high discriminative capability on contact states. For instance, it achieves 96.8\% accuracy in texture recognition and 89.2\% in contact area estimation. This demonstrates that the learned representation effectively captures invariant physical contact mechanics and successfully generalizes to novel visuotactile sensors.

\begin{figure*}[t]
\vspace{2mm}
\centering
\includegraphics[width=\textwidth]{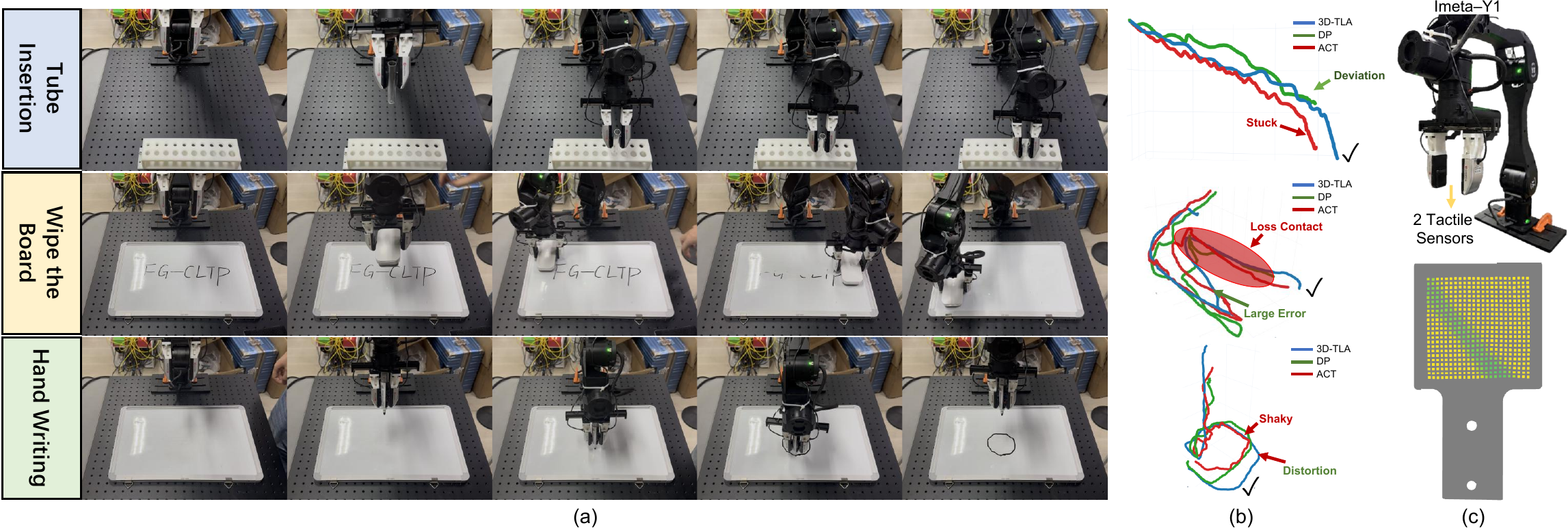}
\caption{Real-world manipulation tasks. (a) Sequential snapshots of the 3D-TLA performing tube insertion, wipe the board, and handwriting. (b) Comparison of end-effector trajectories among 3D-TLA, DP, and ACT. 3D-TLA generates smooth, contact-state-compliant trajectories, whereas the baseline methods suffer from deviations, loss of contact, or distortions. (c) Hardware setup and the collected 3D deformation point cloud.} 
\label{fig:real_world}
\end{figure*}
% -----------------------------------------------------------------------
% PART 3: Real-World VLA
% -----------------------------------------------------------------------
\subsection{Real-World 3D-TLA}

Finally, we validate the performance of our model in real-world contact-rich manipulation scenarios. We designed three challenging tasks that span essential abilities, ranging from in-hand localization and force control to continuous action smoothness. These tasks comprehensively bridge force-sensitive precision control with object manipulation, establishing a holistic benchmark to assess the dynamic tactile perception capabilities of the model in unconstrained environments. Specifically, the first task, tube insertion, requires the robot to insert a tube into a rack under severe visual occlusion, heavily relying on tactile-based in-hand localization. The second task, wipe the board, demands maintaining a uniform contact force and stable surface interaction to clear designated areas effectively. The final task, handwriting, involves writing specific characters on a whiteboard to test fine-grained force control and trajectory precision.

\subsubsection{Real-world Experiment Setup}

% \begin{figure}[!t]
% \centering
% \includegraphics[width=0.5\linewidth]{figures/setup.pdf}
% \caption{Real-World Experiment Setup.} 
% \label{fig:setup}
% \end{figure}

Experiments are conducted using an Imeta Y1 6-DOF robotic arm, two DM-Tac M sensors are mounted to capture high-resolution multimodal tactile signals. All modalities are synchronized at 30 Hz. For each task, we collect 150 expert demonstrations, with trajectory durations ranging from 8s to 30s. 

\subsubsection{Results and Analysis}

In the real-world experiments, we compared 3D-TLA with DP\cite{chi2025diffusion} and ACT\cite{zhao2023learning}, as shown in Fig. \ref{fig:real_world}. Across all evaluated real-world manipulation tasks, our method consistently outperforms ACT and DP, the results are shown in Tab. \ref{tab:exp}. In the tube insertion task, which strictly tests in-hand localization capabilities, 3D-TLA achieves an 85.0\% success rate. This establishes a clear margin over DP (75.0\%) and ACT (70.0\%), demonstrating that our method successfully resolves fine-grained contact states.

%  validating the effective integration of semantically aligned tactile representations with the VLA model.

The advantages of 3D-TLA extend to tasks requiring continuous force control and precise trajectory execution. For the board wiping task, where success demands maintaining a residual uncleaned area of less than 3 cm$^2$, our method achieves a 75.0\% success rate, compared to the 65.0\% attained by both ACT and DP. Furthermore, in the highly sensitive handwriting task, which involves drawing unbroken circles, 3D-TLA reaches a 60.0\% success rate, surpassing DP (50.0\%) and ACT (45.0\%). These consistent performance gains indicate that integrating our fine-grained aligned tactile representation with a VLA model equips the policy to perform precise pose adjustments and handle the dynamics of complex physical interactions more effectively than conventional methods.

\begin{table}[t]
\vspace{2mm}
\caption{Real-World 3D-TLA Evaluation (Metric: \%).}
\centering
\label{tab:exp}
\begin{tabular}{lcccc}
\hline
\textbf{Task}     & \textbf{ACT} & \textbf{DP} & \textbf{3D-TLA (Ours)} \\ \hline
Insert Tube        & 70.0         & 75.0                      & \textbf{85.0 }                  \\
Wipe the Board        & 65.0         & 65.0                      & \textbf{75.0}                   \\
Handwriting              & 45.0         & 50.0                      & \textbf{60.0}                  \\ \hline
\end{tabular}
\end{table}

\section{Conclusion}
% In this work, we presented a novel framework for aligning 3D tactile perception with language semantics to enhance contact-rich manipulation. We introduced the fine-grained contrastive language tactile pretraining (FG-CLTP) method, which bridges the gap between quantitative physical metrics and high-level linguistic semantics. By incorporating discrete numeric tokenization and auxiliary physical regression, our encoder captures both numeric physical attributes and high-level contact semantics, overcoming the limitations of previous qualitatively aligned tactile representations. Furthermore, we developed 3D-TLA, a policy model that integrates our aligned tactile representation with a state-of-the-art VLA backbone. Extensive real-world experiments including tube insertion and blackboard erasing demonstrate that our approach significantly outperforms existing baselines in tasks demanding precise force control and contact-state-aware reasoning. These results validate the necessity of aligning tactile representations in both physical quantities and semantic descriptions for robust robot manipulation.
In this work, we presented a novel framework for aligning 3D tactile perception with language semantics to enhance contact-rich manipulation. We introduced the fine-grained contrastive language tactile pretraining (FG-CLTP) method, which bridges the gap between quantitative physical metrics and high-level linguistic semantics. By incorporating discrete numeric tokenization and auxiliary physical regression, our encoder captures both numeric physical attributes and high-level contact semantics. This approach overcomes the limitations of previous qualitatively aligned representations, reducing the macro average regression MAE by 52.6\% and demonstrating strong cross-sensor generalization with a minimal 3.5\% sim-to-real performance gap due to our tactile 3D point cloud representation. Furthermore, we developed 3D-TLA, a policy model that integrates our aligned tactile representation with a state-of-the-art VLA backbone. Extensive real-world experiments demonstrate that our approach significantly outperforms existing baselines, achieving success rates of 85.0\% in tube insertion and 75.0\% in board wiping. These results validate the necessity of aligning tactile representations in both physical quantities and semantic descriptions to achieve robust manipulation, providing a transferable foundation for the tactile grounding of multimodal robotic models.

%%%%%%%%%%%%%%%%%%%%%%%%%%%%%%%%%%%%%%%%%%%%%%%%%%%%%%%%%%%%%%%%%%%%%%%%%%%%%%%%

%%%%%%%%%%%%%%%%%%%%%%%%%%%%%%%%%%%%%%%%%%%%%%%%%%%%%%%%%%%%%%%%%%%%%%%%%%%%%%%%



\section*{ACKNOWLEDGMENT}
We would like to extend our thanks to Daimon Robotics for their support with tactile sensing devices.
% The preferred spelling of the word ÒacknowledgmentÓ in America is without an ÒeÓ after the ÒgÓ. Avoid the stilted expression, ÒOne of us (R. B. G.) thanks . . .Ó  Instead, try ÒR. B. G. thanksÓ. Put sponsor acknowledgments in the unnumbered footnote on the first page.

\bibliographystyle{IEEEtran}
\bibliography{example}

\end{document}